\documentclass[9pt,conference]{IEEEtran}
\title{Fitting tree model with CNN and geodesics to track vessels and application to Ultrasound Localization Microscopy data}
\author{\IEEEauthorblockN{Théo Bertrand}
\IEEEauthorblockA{CEREMADE, UMR CNRS 7534,\\
University Paris Dauphine, PSL, France\\
Email: tbertrand@ceremade.dauphine.fr}
\and
\IEEEauthorblockN{Laurent D. Cohen}
\IEEEauthorblockA{CEREMADE, UMR CNRS 7534,\\
University Paris Dauphine, PSL, France\\
Email: cohen@ceremade.dauphine.fr}
}

\date{November 2023}

\usepackage{graphicx}
\usepackage{hyperref}
\usepackage{tabularx}

\usepackage[english]{babel}

\usepackage{amsmath}
\usepackage{amsfonts}
\usepackage{amssymb}

\usepackage{enumitem}

\usepackage{tikz}
\usepackage{adjustbox}
\usepackage{subcaption}

\usepackage{ bbold }

\newcommand{\ud}{\text{d}}
\newcommand{\R}{\mathbb{R}}

\begin{document}

\maketitle

\begin{abstract}

Segmentation of tubular structures in vascular imaging is a well studied task, although it is rare that we try to infuse knowledge of the tree-like structure of the regions to be detected. Our work focuses on detecting the important landmarks in the vascular network (via CNN performing both localization and classification of the points of interest) and representing vessels as the edges in some minimal distance tree graph. We leverage geodesic methods relevant to the detection of vessels and their geometry, making use of the space of positions and orientations so that 2D vessels can be accurately represented as trees. We build our model to carry tracking on Ultrasound Localization Microscopy (ULM) data, proposing to build a good cost function for tracking on this type of data. We also test our framework on synthetic and eye fundus data. Results show that scarcity of well annotated ULM data is an obstacle to localization of vascular landmarks but the Orientation Score built from ULM data yields good geodesics for tracking blood vessels. 
\end{abstract}

\section{Introduction}

Ultrasound Localization Microscopy is a quite recent imaging technique that allows users to bypass the compromise between precision and depth of penetration in ultrasound imaging.

It allows one to make highly resolved images of the vascular network deeper in the skin tissues with the help of micro bubbles used as contrast agents. We refer to \cite{couture-ultrasound-2018} for an overview of the super resolution method.

\begin{figure}
    \centering
    \includegraphics[trim={1cm 0cm 1cm .5cm},clip,width=0.9\linewidth]{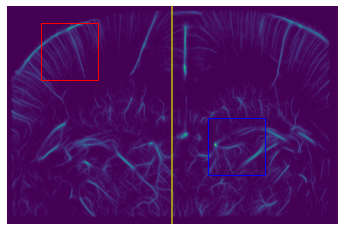}
    \includegraphics[trim={3cm 1cm 3cm 1cm},clip,width=0.45\linewidth]{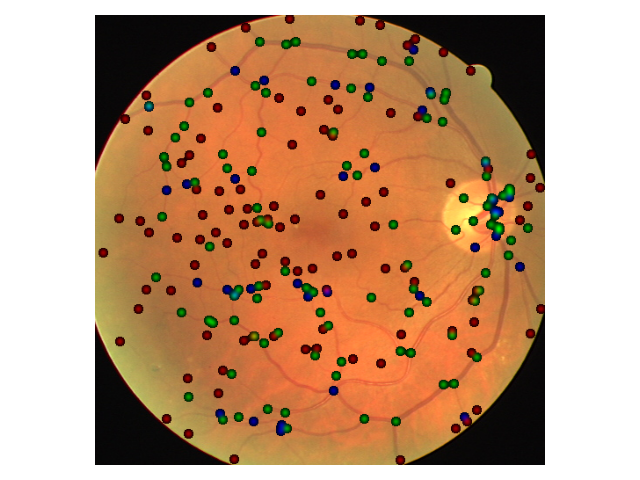}
    \includegraphics[trim={3cm 1cm 3cm 1cm},clip,width=0.45\linewidth]{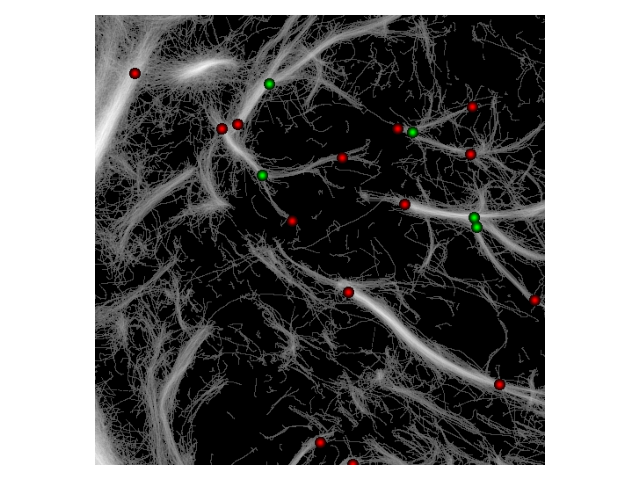}
    \caption{Top : Patches are made from a high resolution image by cropping patches taken uniformly from each brain half. Left : eye fundus image overlayed with heatmap of vascular landmarks. Right : ULM image overlayed with heatmap of vascular landmarks.}
    \label{fig:data_crop}
\end{figure}

In the present work, we introduce a new workflow for complete end-to-end detection of vascular structures on ULM images, using deep learning to detect landmarks (see Figure \ref{fig:data_crop}) allowing tracking of vessels as edges in a tree graph with landmarks as vertices. Our approach differs from classical Perceptual Grouping for blood vessel tracking as performed in \cite{bekkers_nilpotent_2018,benmansour_fast_2009}. Indeed we are trying to take advantage of long geodesics tracking blood vessels across an image, that should behave well given the amount of literature on the subject. While Perceptual Grouping usually focuses on computing short geodesics between close points spread across the vessel network, we aim to compute few long geodesics between key landmakrs of the vasculature.
We try to take advandatage of the information specifically given by ULM imaging, but it must be noted that it is possible to adapt on other types of images, for instance eye fundus images obtained via direct photography. To do so, one may need to evaluate local orientation information as we will see in section 3. From a 2D image it can be done using Orientation Scores \cite{duits2007OS} or similar transforms such as the ones presented in \cite{OS-jiong} to lift a 2D image set in the plane to the 3-dimensional space of positions and orientations.
Raw ULM data consists in ultrasound signals for imaging blood vessels indirectly. Indeed, instead of directly viewing the response of the tissues, it is the non-linear response of microbubbles used as contrast agents that is recovered and then treated to recover the position of the bubbles. The set of the position of the microbubbles transported in the blood vessels allows one to recover a highly-resolved image of the vascular tree by projecting back on a grid as fine as needed (although limited by the number of detected microbubbles). Some methods (see for instance \cite{Pengfei_tracking_2018}) then use the detected points to try and recover trajectories of microbubbles along successive images, thus allowing one to interpolate along those trajectories and infer even more points in order to provide finer images.  The data available and used in this work is composed of detected trajectories of microbubbles. We want to exploit previously recomposed trajectories by taking into account the additional information contained in said trajectories, which are not only point clouds but have velocity and temporal information.\\



Detection, segmentation or tracking tasks on medical images are widely studied problems. 

The contributions of our work include :
\begin{itemize}[noitemsep,topsep=0pt]
    \item working with ULM data, defining a Riemannian metric in order to track vessels in ULM images,
    \item dealing with scarcity of data : 2 different high resolution images to make both the training and validation set, 
    \item carrying out detection of vascular landmarks in such context,
    \item fitting a tree model with geodesics as edges to take into account geometric and topological aspects into the tracking, thus investigating the efficiency of using the tree-like nature of vasculature to perform the tracking,
    \item comparing results on synthetic data (hand-made black and white images to fit the framework used for ULM data, i.e. few big images) and eye fundus images (more images, but smaller).
\end{itemize}
    

Vessel segmentation is usually performed by computing scores of vesselness on the image, see the seminal work \cite{frangi-multiscale-1998} and more recent work \cite{jerman-enhancement-2016}. The main idea in these works being that high vesselness corresponds to regions in the image where one orientation is dominant.
Vesselness is then defined as a function of the eigenvalues of the Hessian (in dimension 2, one eigenvalue being significantly higher than the other indicates a tubular region). Modern methods of transposing the image in a higher dimensional setting of Position-Orientation space were used in \cite{OS-jiong}.

Other methods of vascular segmentation include machine and deep learning methods that have become accessible thanks to the availability of annotated data. We can cite for instance \cite{Oliveira-2018} that uses a fully-convolutional U-net for segmentation task on eye fundus images.

A few works have already approached the problem of localizing vascular landmarks in eye fundus images \cite{abbasi2015IOSTAR,pratt_automatic_2017, wang_automatic_2023,calvo_automatic_2011,tetteh_deepvesselnet_2020} but they usually first focus on providing a segmentation mask of the blood vessels in the image before carrying post-processing on the segmentation to infer the positions of the landmarks (\emph{endpoints}, \emph{crossings} of \emph{bifurcations} of blood vessels). \cite{hervella} tries to tackle the problem of finding landmarks directly from input data, and we will be building on this method here.

The second part of our workflow makes use of geodesic curves to track vascular structures in the input images. Tracking vascular structures using geodesics has been done multiple times for instance in \cite{deschamp-cohen-2001,fethallah-cohen-2011}, those methods have had multiple extensions, for instance taking advantage of roto-translation group \cite{Bekkers-2014} or adding vessel width information \cite{li2007vessels}.

\section{Detecting vascular landmarks}
\label{Detection}
The approach to detect the vascular landmarks is very much like the one used in \cite{hervella}. 
The novelty of our approach is the scarcity of ULM data we use for learning, the integration of \emph{endpoints} in the detection task, and the tracking described further down.

Indeed, we want to generate heatmaps with multiple channels indicating probable locations of vascular landmarks. 

We train a single U-net architecture to learn the localization and classification of interesting points in a 2D ULM image of brain vessels.
The network outputs 4 channels : 3 for different types of landmarks (\emph{endpoints}, \emph{bifurcations}, \emph{crossings}) and a last one to relax classification. The heatmap predicted by the CNN is then filtered to get the position of local maxima (after thresholding at level $r$ of output of network to reduce noise).


\subsection{ULM Data}
Our data is composed of few highly resolved images of rat brains obtained via ULM imaging. The scarcity of data is a usual problem in medical images processing.
We were provided with two such images of rat brains (two similar plans were imaged) by \cite{ULM-data}, we then proceeded to uniformly cut those big ($3210\times 2675$) images into smaller square patches. This way, we aggregate around 42 ULM images, making a training dataset of $21$ and another set for validation of $21$ images. We make sure that there are patches from both original images in both sets. We also make sure that there is no overlapping between the training and validation datasets by using different brain halves to make those patches (see Figure \ref{fig:data_crop} top).

As there is no available dataset annotation for segmentation task of ULM data nor for the landmark localization and classification task, annotation for the latter was produced by one of the authors. The dataset annotation was made by selecting the point landmarks by hand with the appropriate tool \cite{make-sense}. One great difficulty of our approach is that we are highly dependent on the accuracy of the initial annotation of the data which can be hard given that there are multiple visible vessel sizes on ULM images.

To prove the efficency of our method, we also apply it to two other datasets : one that consists in two synthetic images of tubular structures arranged into a network, for training and validation, they are very big and are used to imitate the case of ULM images (high resolution, few images, thus cut into smaller patches); the other one is simply a dataset using both images and groundtruth from the DRIVE and IOSTAR datasets, much like in the previously cited work \cite{hervella}.

\subsection{Training}
The training loss is defined as $\mathcal{L}(\theta,x,\hat{y}) = \|f_{\theta}(x) - \hat{y}\|_2^2$ the mean squared error (MSE), where $\hat{y}$ is the position of the labeled features in the input images in our dataset convolved with a gaussian kernel $\hat{y} = \sum_{y\in D} k_{\sigma}*\delta_y$, $f_{\theta}$ is our CNN architecture with parameters $\theta$, applied on the $x$ input image. $k_{\sigma}$ is a gaussian kernel with chosen standard $\sigma$.

Going through our data we minimize the loss evaluated on the training set over the space of parameters $\theta \in \Theta$. We may recall that the U-net architecture is an encoder-decoder architecture composed of multiple convolution layers ($3\times 3$ filters and leaky ReLU activation) and with skip-connections. \cite{Unet} is the fundamental work introducing this architecture.

To make up for the small size of our dataset, we perform data augmentation via horizontal symmetries, translations, rotations, all randomly applied with predefined parameters. It allows us to artificially expand our training dataset, leveraging equivariance of our task by the action of those transformations.





\section{Finding appropriate geodesics}
Geodesics have been used for tracking vessels in vascular images for a long time now. The works \cite{deschamp-cohen-2001,fethallah-cohen-2011} laid good basis for such work, and the book \cite{peyre-geodesic-nodate} that is a good introduction to the use of geodesics for image analysis. These works leverage our knowledge of geodesic curves and numerical algorithms allowing us to compute them to track tubular structures on medical images.

\subsection{Geodesics for vessel tracking}
\label{sec:model_RS}

Geodesics are curves that minimize a given energy $\mathcal{E}(\gamma) = \int_0^1 \mathcal{P}\left(\gamma(s), \gamma'(s)\right) \ud s$ with $\gamma \in \text{Lip}([0,1], \Omega),$  where $\mathcal{P}:\Omega \times \R^d \longrightarrow \R_+$ is some given feature potential.

In fact, more than an energy, with a few hypothesis on $\mathcal{P}$, $\mathcal{E}$ may be seen as the length of the curve $\gamma$ in some geometry described by the potential $\mathcal{P}$. 

Given two points $x_0,x_1 \in \R^d$, a curve minimizing the energy $\mathcal{E}$ under the constraints $\gamma(0)=x_0$ and $\gamma(1) = x_1$ is called a \textit{geodesic} or \textit{minimal path} (joining $x_0$ and $x_1$) according to the metric defined by $\mathcal{P}$.

In general we will limit ourselves to the cases where $\mathcal{P}$ is a Riemannian metric, i.e. the square root of a quadratic form associated with a positive definite tensor field $M$ : $\mathcal{P}(x,v) = \sqrt{\left<M(x)v,v\right>}$. It can be interpreted as a local measure of the norm of some velocity vector $v$ in the neighbourhood of $x$.



The function $\mathcal{E}$ then defines a distance map : 
$d(x,y) = \inf_{ \gamma(0)=x, \gamma(1) = y.} \mathcal{E}(\gamma).$


 
 
 


Vessel tracking can be performed by finding geodesics on an image of the vessels. To do so, we simply need to define a metric that is well adapted. 

For now, we will restrict ourselves to Riemmanian metrics defined on the homogenous space of positions and orientations $\mathbb{M}_d = \R^d \times \mathbb{P}^{d-1}$ with $\mathbb{P}^{d-1} \simeq\mathbb{S}^{d-1}/\{-1,1\}$, $\mathbb{P}^{d-1}$ allows us to assimilate features that have the same direction but not the same sign. we will use the relaxed Reeds-Shepp metric that is well-studied, Riemannian and penalizes curves that are not planar.

The relaxed Reeds-Shepp metric is the one associated with the metric tensor defined by :
\[\mathcal{P}_{\varepsilon}((x, \theta),(\Dot{x}, \Dot{\theta}))^2 =  C((x,\theta))^2 (|\Dot{x}\cdot e_{\theta}|^2 + \frac{1}{\varepsilon^2}|\Dot{x}\wedge e_{\theta}|^2 + \xi^2|\Dot{\theta}|^2),\]
with  $\xi, \varepsilon \in \R,$ $e_{\theta}$ the unit vector with orientation $\theta$. $C$ is a cost function, in the following it will be defined as $C = \frac{1}{1+\lambda W^2}$ with $\lambda =10^{3}$ and $W$ a $[0,1]$-valued score built from the image.

This vesselness score $W$ is important because it allows us to associate a $\theta$ coordinate to all the detected landmarks by finding the orientation $\theta$ maximizing the score at its position.

We also look to enrich our orientation-dependant score by adding information from the detection of vascular landmarks by imposing $\forall \theta, W(x,\theta)=1$ if a \emph{bifurcation} has been detected at position $x,$ so that the landmark point is accessible from any orientation.

Similarly, if a landmark has been found and classified as a \emph{crossing}, we add a new point to our set of detected points located at this position but with second maximum intensity in the vesselness score.

Such geodesics are well-studied and have already been used in previous works to accurately track blood vessels (for instance in \cite{duits2018optimal}). The main asset of this model is that it helps avoid shortcuts in the case where two different vessels cross in a 2D image. 






The geodesic distance can be computed efficiently and fast using the Fast Marching Algorithm, we refer to \cite{duits2018optimal} and the attached library for efficient computational tools used in the present work.

 \subsection{Clustering landmarks using the geodesic graph}
 
 Once we have defined a proper geodesic distance and we are able to effectively compute it, we can build a matrix $D = (d(x_i,x_j))_{1\leq i,j \leq n_l}$ of pairwise distance, where the $x_i$ are the $n_l$ detected landmarks. 
 
 This step is the computational bottleneck as it requires to compute $n_l(n_l-1)/2$ coefficients, meaning solving $n_l$ times the Fast Marching algorithm to iteratively fill the lines of the matrix, computing the map $d(x_i,\cdot)$ at every iteration $i$. Thus the complexity is around $\mathcal{O} (n_l N \log(N))$ with $N$ the number of points. 
 
 The computed pairwise distance matrix thus defines a complete weighted graph that we call the geodesic graph. On a single image there may be many different groups of vessels that appear, with the computation of the pairwise distance we have already computed the geodesic curves between each pair of points. We then need to keep only the groups of points that are relevant for representation of the vessels. To cut the complete graph into smaller connected components, we will simply perform hierarchical clustering on the graph. Indeed, if the metric is well chosen to make landmarks linked by the vascular network near for the distance $d$, and landmarks not connected by the vascular network far, we simply group aggregate points that are near and separate them from the others under some condition of threshold distance $s_{cluster}$.

We may cite the work \cite{muller-hierarchical} as a reliable source for theory and algorithms for hierarchical clustering.

\subsection{Linking landmarks through the geodesic graph}

Our landmarks are now separated into multiple groups. Each of these group supposedly represents one connected component of the visible vascular network in the 2D image.

Those smaller groups represent smaller complete graphs, but we still need to select which of the computed geodesics represent the vessels.

Now the objective is to only keep the curves that accurately represent blood vessels in the image. This can be done by removing some of the edges in each smaller graph.

We need a few properties for the target graph, inferred from the idea we have of the representation of blood vessels :
\begin{itemize}[noitemsep,topsep=0pt]
    \item As it should represent vessels, it needs to be "$1-$dimensional" i.e. it is represented by a planar graph.
    \item It does not have cycles.
    \item It is small for the geodesic distance (if previously well chosen metric).
\end{itemize}

A good heuristic to have these properties is to look for the minimal spanning tree in each smaller complete graphs.

A few algorithms are available to perform this computation efficiently. For Kruskal's algorithm, time complexity is of the order of the sorting of the edges' weights $\mathcal{O}(e \log(e))$ with $e$ the number of edges in the graph. 

\section{Results and Discussion}


\subsection{Synthetic data}

To test our framework we can execute our algorithm on synthetic hand drawn data. 
The main difference is that if we use 2D images to simulate our network, there is no straightforward way to define orientation-lifted images as we do with ULM microbubble trajectories data. To generate such orientation-lifted images, we leverage our knowledge of Orientation score techniques.


We define a simple transformation by convolution with the help of anisotropic gaussian kernels : the anisotropy in a direction $\theta$ of the kernel will allow us to select only the parts where the local features are aligned with this direction.

This kernel defines the lifting operator $\Phi$ : 
\begin{align*}
        \forall u \in L^2(\Omega), &\quad \forall (x,\theta) \in \Omega \times [0,\pi[,\\
        &u^{\text{lifted}}(x, \theta) = (\Phi u)(x,\theta) = (k_{\theta}^{\text{lift}} * u)(x),
\end{align*}



To train the CNN to detect landmarks, we use two synthetic images : one for the training set and the other for the validation and apply the same approach as in the case of ULM data.

\begin{figure}
    \centering
    \includegraphics[trim={2.5cm .5cm 2.5cm .5cm},clip,width = 0.45 \linewidth]{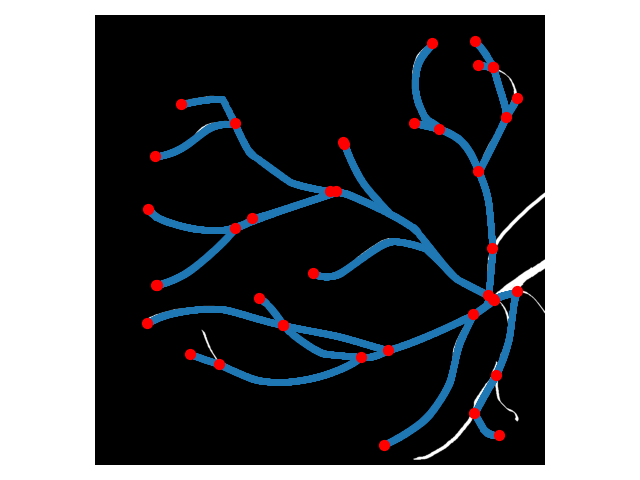}
    \includegraphics[trim={2.5cm 0cm 18.5cm 0cm},clip,width = 0.45 \linewidth]{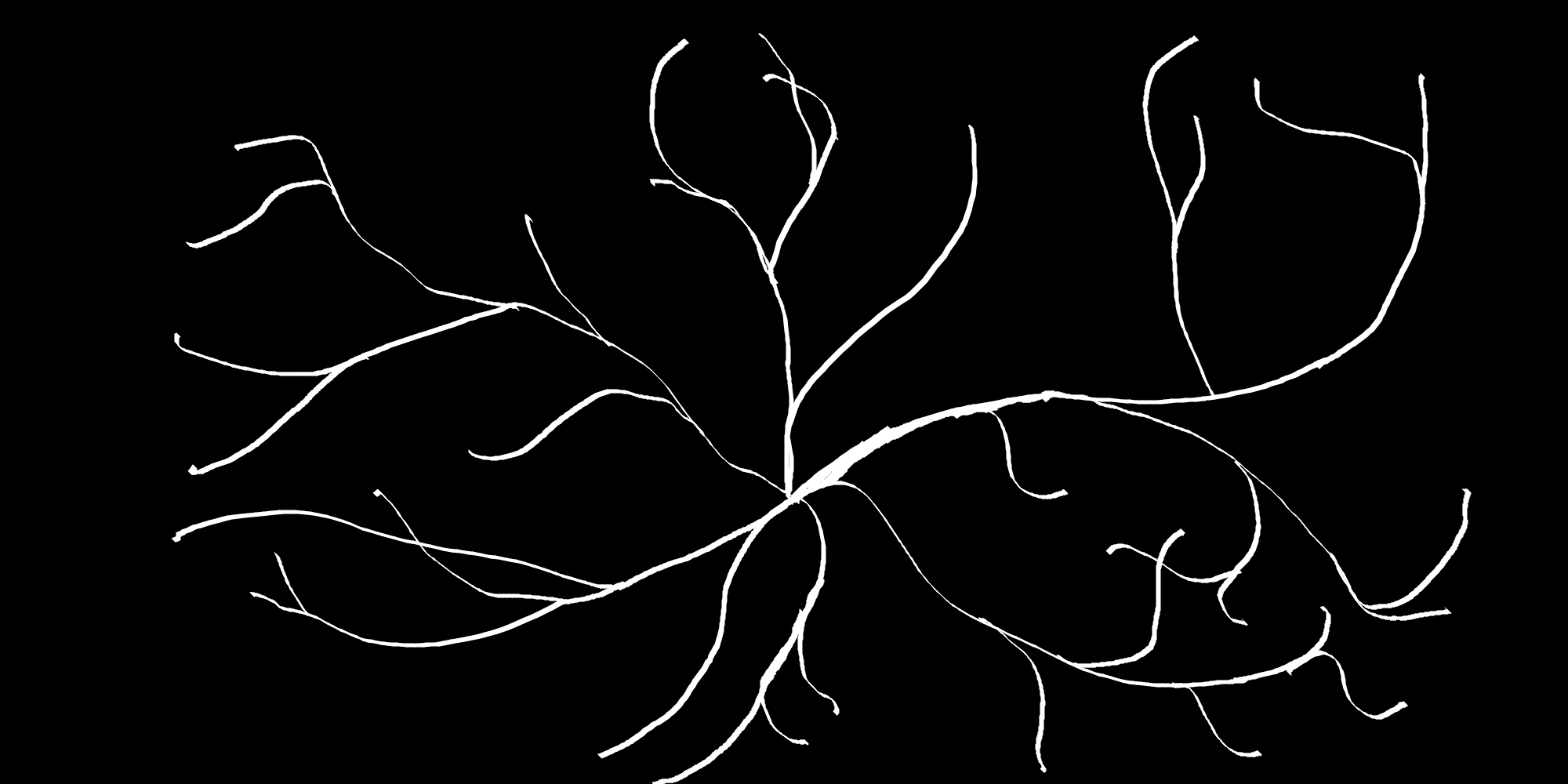}
    \caption{Geodesic graph on half of the synthetic validation image, with $N_{\theta} = 128$. Left image shows detected landmarks points and the geodesics linking them, right image shows the input image.}
    \label{fig:Synthetic_NetOutput}
\end{figure}
\vspace{.1cm}


\textbf{Results :} At the landmark-detection task level, we were able to attain an F1 score of about 77\% on the validation dataset with hyper-parameters selected by hand and \textit{recall} scores of 80\% and 75\% respectively) as evaluated on the whole validation image.
During the training process, validation is made on a set of images taken as random crops from the original big image. The validation scores obtained on those smaller ($256\times 256$) patches tend to give similar scores on average along the dataset, although the scores oscillate a lot along epochs. Figure \ref{fig:Synthetic_NetOutput} shows the resulting tracking of synthetic vessels and the corresponding geodesic graph.

\subsection{Eye fundus image}

To reinforce our methodology, we apply our workflow to a classical dataset of eye fundus images, as a middle ground between synthetic and ULM data.

The data came from the IOSTAR and DRIVE dataset \cite{abbasi2015IOSTAR}, it was split into a training set (30 images), a validation set (10 images) and a test set (21 images). 

For the training, we perform random data augmentation with translations and rotations, as to avoid overfitting and take advantage of equivariance properties of the task at hand, and also random crops of fixed size. 

We built the vesselness score $W$ for the computation of minimal paths by first applying a classical Frangi filter \cite{frangi-multiscale-1998} on the input images and then lifting the filtered images via Orientation Score as described in the previous subsection.

\begin{figure}
    \centering
    \includegraphics[trim={2.5cm .5cm 2.5cm .5cm},clip,width = 0.45 \linewidth]{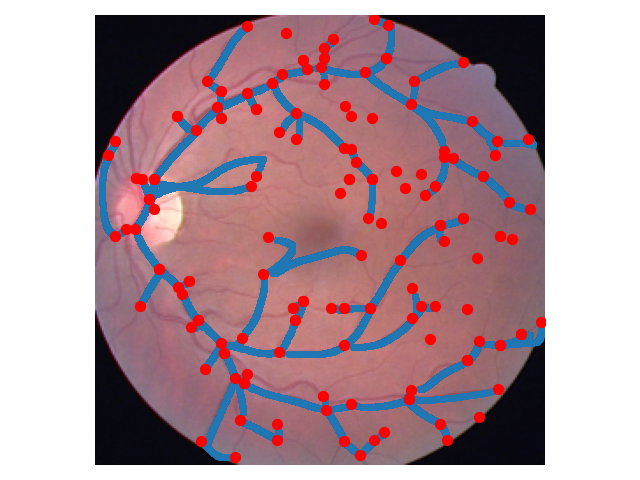}
    \includegraphics[trim={.2cm .2cm .2cm .2cm}, clip, width = 0.45 \linewidth]{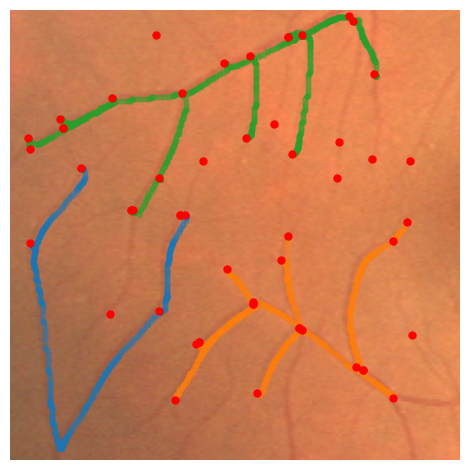}
    \caption{Geodesic tracking performed on two validation images from the DRIVE and IOSTAR dataset, with $N_{\theta} = 64.$ Big red points are the detected landmarks and curves are the selected tree structures.}
    \label{fig:IOSTAR_NetOutput}
\end{figure}

\textbf{Results :} We were able to achieve satisifying results of about 60\% in F1 score on both validation dataset and test dataset (after hyper-parameter searching on the validation data). These results on the landmarks detection task is not as good as the ones presented in \cite{hervella} but they predict only two classes of landmarks (\emph{crossings} and \emph{bifurcations}) whereas we also added the additional \emph{endpoint} class. Figure \ref{fig:IOSTAR_NetOutput} shows a sample of our tracking performed on eye fundus data.

\subsection{Rat Brain ULM data}

The main ideas of the data processing for ULM data has already been described in Section \ref{Detection}.

We want to make full use of ULM data and use the initial set of microbubbles path from the available data \cite{ULM-data} to construct the cost function $C$ in the relaxed Reeds-Shepp model as detailed in Section \ref{sec:model_RS}. With this goal in mind we define $W$ by building directly the Orientation score from the histogram of microbubbles in the dataset just like it is done for the input $2D$ image, but this time we add the orientation of the given velocity vector for the orientation coordinate. After renormalization it gives us a function $W(x,\theta)$ with values beteween $0$ and $1.$

\begin{figure}
    \centering
    \includegraphics[trim={2.5cm .5cm 2.5cm .5cm},clip,width = 0.45 \linewidth]{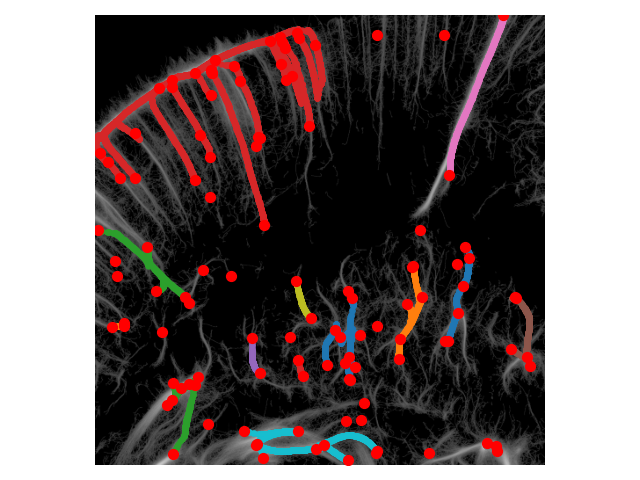}
    \includegraphics[trim={2.5cm .5cm 2.5cm .5cm},clip,width = 0.45 \linewidth]{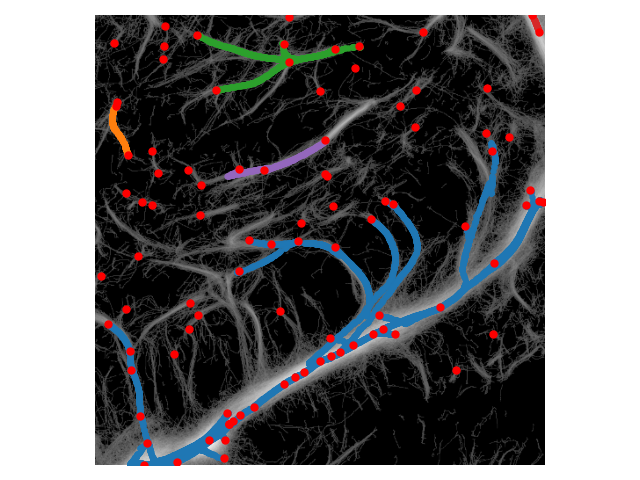}
    \caption{Geodesic graph on patches cropped from the ULM validation dataset (taken respectively from left and right parts of rat brain), with $N_{\theta}=64$. Big red points are the detected landmarks and curves are the selected tree structures.}
    \label{fig:ULM_NetOutput}
\end{figure}

\textbf{Results :}
With the described approach to learning the detection of landmarks, we were only able to reach low mean F1 scores of around 20 \% (computed on $512\times 512$ patches). Even with such a low score on the detection-classification task, we are able to track a few of the vessels in the image, as shown in Figure \ref{fig:ULM_NetOutput}. Still, some big vessels remain untracked because some points were not found at their tips.

\subsection{Discussion}

Application to real world data does not seem to work in a very satisfying way. 

We may note the following behaviours observed after training with different hyper-parameters :

\begin{itemize}
    \item The recovery of the geodesic tree structure is highly sensitive to change in hyper-parameters (in the definition of the metric tensor or the dependance on the position of the detected points).
    \item Our framework is thought for ULM images and does not necessarily adapt well to the eye fundus images dataset considered in the tracking step, although results might get better if one can tune the Orientation Score well enough such that orientation are well separated and landmarks can be linked i.e. reasonably close for the geodesic distance.
    \item Defining the Orientation-dependent cost function from the position of the microbubbles and their estimated velocity vector seems to be a good approach to perform tracking on ULM data as we can see from recovered geodesics in Figure \ref{fig:ULM_NetOutput}
    \item The results on the synthetic images tend to show that if we can provide a good enough segmentation it would be relatively easy to provide a good detection of landmarks and retrieve a good geodesic tree tracking.
    \end{itemize}





\section{Conclusion and further works}
In this work we have investigated the possibility to recover a complete tracking of the vessels in 2D images of vascular networks. It was done using CNN techniques from the literature to extract vascular landmarks that define the main points of interest defining the network. Our method is interesting because it fits a length-minimizing tree model to the image (using geodesics in a certain geometry to represent vessels) and thus includes both topological (tree-like structure) and geometrical (fitting geodesics) information to our tracking.

Although results on real world data are not satisfying for a complete recovery of the vasculature, we have shown the potential of using ULM data and the information they carry can be used to accurately track vessels.

Further research prospects include incorporating scale information or scale equivariance to distinguish vessels and help the localization process and also provide width information  
\pagebreak
\section*{Acknowledgements}
This work was funded in part by the French government under management of Agence Nationale de la Recherche as part
of the "Investissements d’avenir" program, reference ANR-19-P3IA-0001 (PRAIRIE 3IA Institute).
The authors would to thank Dr Olivier Couture and his team for the support on ULM data, and  Dr Erik Bekkers and Dr Jiong Zhang for the access to the DRIVE and IOSTAR datasets.

\section*{Conflict of Interest Disclosure}
No conflict of interest to be declared.

\section*{Compliance with Ethical Standards}
The Numerical study in this work required no ethical approval in itself.

ULM data were made available in open access in \cite{ULM-data}. Ethical approval was not required as confirmed by the declared approval of the Ethics Comittee on Animal Experiments of Paris Centre et Sud and Paris Descartes.

DRIVE and IOSTAR data come from \cite{staal2004ridge} and \cite{abbasi2015IOSTAR} available in open access.

\bibliographystyle{IEEEtran.bst}
\bibliography{references}

\end{document}